\documentclass[11pt]{article}

\usepackage[margin=1.1in]{geometry}
\usepackage[T1]{fontenc}
\usepackage[utf8]{inputenc}
\usepackage{lmodern}
\usepackage{microtype}
\usepackage{amsmath,amssymb}
\usepackage{booktabs}
\usepackage{graphicx}
\usepackage{caption}
\usepackage{subcaption}
\usepackage[numbers,sort&compress]{natbib}
\usepackage{xcolor}
\usepackage[colorlinks=true,linkcolor=blue!60!black,citecolor=blue!60!black,urlcolor=blue!60!black]{hyperref}
\usepackage{enumitem}
\usepackage{array}

\newcommand{\cErrTarget}{0.10}
\newcommand{\cRetained}{0.97}
\newcommand{\cSeed}{20261001}
\newcommand{\cBootN}{10{,}000}
\newcommand{\cBhQ}{0.05}
\newcommand{\cHeadsN}{10{,}000}
\newcommand{\cWinClm}{2048}
\newcommand{\cWinLaya}{512}
\newcommand{\cWinGen}{4096}
\newcommand{\cWinRm}{4096}
\newcommand{\truncClmRb}{0.01}
\newcommand{\keptClmRb}{0.99}
\newcommand{\truncClmRbb}{0.56}
\newcommand{\keptClmRbb}{0.83}
\newcommand{\truncClmRbp}{0.06}
\newcommand{\keptClmRbp}{1.00}
\newcommand{\truncClmRmb}{0.05}
\newcommand{\keptClmRmb}{0.99}
\newcommand{\truncClmJb}{0.15}
\newcommand{\keptClmJb}{0.97}
\newcommand{\truncClmHalu}{0.19}
\newcommand{\keptClmHalu}{0.71}
\newcommand{\truncGenRb}{0.00}
\newcommand{\keptGenRb}{1.00}
\newcommand{\truncGenRbb}{0.09}
\newcommand{\keptGenRbb}{1.00}
\newcommand{\truncGenRbp}{0.00}
\newcommand{\keptGenRbp}{1.00}
\newcommand{\truncGenRmb}{0.01}
\newcommand{\keptGenRmb}{1.00}
\newcommand{\truncGenJb}{0.06}
\newcommand{\keptGenJb}{0.99}
\newcommand{\truncGenHalu}{0.19}
\newcommand{\keptGenHalu}{0.71}
\newcommand{\truncLayaRb}{0.40}
\newcommand{\keptLayaRb}{0.76}
\newcommand{\truncLayaRbb}{0.93}
\newcommand{\keptLayaRbb}{0.27}
\newcommand{\truncLayaRbp}{0.77}
\newcommand{\keptLayaRbp}{0.50}
\newcommand{\truncLayaRmb}{0.84}
\newcommand{\keptLayaRmb}{0.51}
\newcommand{\truncLayaJb}{0.99}
\newcommand{\keptLayaJb}{0.42}
\newcommand{\truncLayaHalu}{0.25}
\newcommand{\keptLayaHalu}{0.54}
\newcommand{\truncRmRb}{0.00}
\newcommand{\keptRmRb}{1.00}
\newcommand{\truncRmRbb}{0.06}
\newcommand{\keptRmRbb}{1.00}
\newcommand{\truncRmRbp}{0.00}
\newcommand{\keptRmRbp}{1.00}
\newcommand{\truncRmRmb}{0.00}
\newcommand{\keptRmRmb}{1.00}
\newcommand{\truncRmJb}{0.06}
\newcommand{\keptRmJb}{0.99}
\newcommand{\truncRmHalu}{0.19}
\newcommand{\keptRmHalu}{0.71}
\newcommand{\latClmRb}{338}
\newcommand{\latClmRbb}{674}
\newcommand{\latClmRbp}{233}
\newcommand{\latClmRmb}{106}
\newcommand{\latClmJb}{292}
\newcommand{\latClmHalu}{63}
\newcommand{\latGencRb}{746}
\newcommand{\latGencRbb}{1{,}058}
\newcommand{\latGencRbp}{898}
\newcommand{\latGencRmb}{780}
\newcommand{\latGencJb}{937}
\newcommand{\latGencHalu}{635}
\newcommand{\latLayaRb}{23}
\newcommand{\latLayaRbb}{29}
\newcommand{\latLayaRbp}{24}
\newcommand{\latLayaRmb}{22}
\newcommand{\latLayaJb}{32}
\newcommand{\latLayaHalu}{23}
\newcommand{\latRmRb}{244}
\newcommand{\latRmRbb}{558}
\newcommand{\latRmRbp}{359}
\newcommand{\latRmRmb}{230}
\newcommand{\latRmJb}{331}
\newcommand{\accClmRb}{0.549}
\newcommand{\accLoClmRb}{0.527}
\newcommand{\accHiClmRb}{0.570}
\newcommand{\accLayaRb}{0.512}
\newcommand{\accGencRb}{0.778}
\newcommand{\accRmRb}{0.976}
\newcommand{\accTfirstRb}{0.500}
\newcommand{\accTrandRb}{0.500}
\newcommand{\accTlenRb}{0.441}
\newcommand{\accGensRb}{0.868}
\newcommand{\accClmftRb}{0.521}
\newcommand{\accClmRbb}{0.351}
\newcommand{\accLayaRbb}{0.280}
\newcommand{\accGencRbb}{0.675}
\newcommand{\accRmRbb}{0.876}
\newcommand{\accTfirstRbb}{0.263}
\newcommand{\accTrandRbb}{0.250}
\newcommand{\accTlenRbb}{0.223}
\newcommand{\accGensRbb}{0.732}
\newcommand{\accClmftRbb}{0.387}
\newcommand{\accClmRbp}{0.593}
\newcommand{\accLayaRbp}{0.526}
\newcommand{\accGencRbp}{0.758}
\newcommand{\accRmRbp}{0.950}
\newcommand{\accTfirstRbp}{0.500}
\newcommand{\accTrandRbp}{0.500}
\newcommand{\accTlenRbp}{0.418}
\newcommand{\accGensRbp}{0.827}
\newcommand{\accClmftRbp}{0.647}
\newcommand{\accClmRmb}{0.508}
\newcommand{\accLoClmRmb}{0.487}
\newcommand{\accHiClmRmb}{0.529}
\newcommand{\accLayaRmb}{0.494}
\newcommand{\accGencRmb}{0.676}
\newcommand{\accRmRmb}{0.923}
\newcommand{\accTfirstRmb}{0.500}
\newcommand{\accTrandRmb}{0.500}
\newcommand{\accTlenRmb}{0.468}
\newcommand{\accGensRmb}{0.726}
\newcommand{\accClmftRmb}{0.539}
\newcommand{\accClmJb}{0.475}
\newcommand{\accLoClmJb}{0.426}
\newcommand{\accHiClmJb}{0.518}
\newcommand{\accLayaJb}{0.498}
\newcommand{\accGencJb}{0.611}
\newcommand{\accRmJb}{0.764}
\newcommand{\accTfirstJb}{0.500}
\newcommand{\accTrandJb}{0.500}
\newcommand{\accTlenJb}{0.467}
\newcommand{\accGensJb}{0.615}
\newcommand{\accClmftJb}{0.535}
\newcommand{\accClmHalu}{0.581}
\newcommand{\accLayaHalu}{0.544}
\newcommand{\accGencHalu}{0.728}
\newcommand{\accTfirstHalu}{0.581}
\newcommand{\accTrandHalu}{0.501}
\newcommand{\accTlenHalu}{0.425}
\newcommand{\accGensHalu}{0.745}
\newcommand{\accClmftHalu}{0.581}
\newcommand{\headsValAcc}{0.782}
\newcommand{\headsTestAcc}{0.764}
\newcommand{\headsInitTest}{0.623}
\newcommand{\calTsingle}{20.1}
\newcommand{\calEceIdRb}{0.235}
\newcommand{\calGapIdRb}{+0.235}
\newcommand{\calEceTsRb}{0.032}
\newcommand{\calEceTbRb}{0.032}
\newcommand{\calEceIsoRb}{0.015}
\newcommand{\calEceIdRbb}{0.402}
\newcommand{\calGapIdRbb}{+0.401}
\newcommand{\calEceTsRbb}{0.061}
\newcommand{\calEceTbRbb}{0.023}
\newcommand{\calEceIsoRbb}{0.060}
\newcommand{\calEceIdRbp}{0.283}
\newcommand{\calGapIdRbp}{+0.283}
\newcommand{\calEceTsRbp}{0.049}
\newcommand{\calEceTbRbp}{0.027}
\newcommand{\calEceIsoRbp}{0.020}
\newcommand{\calEceIdRmb}{0.343}
\newcommand{\calGapIdRmb}{+0.343}
\newcommand{\calEceTsRmb}{0.042}
\newcommand{\calEceTbRmb}{0.042}
\newcommand{\calEceIsoRmb}{0.028}
\newcommand{\calEceIdJb}{0.354}
\newcommand{\calGapIdJb}{+0.351}
\newcommand{\calEceTsJb}{0.062}
\newcommand{\calEceTbJb}{0.083}
\newcommand{\calEceIsoJb}{0.122}
\newcommand{\calEceIdHalu}{0.255}
\newcommand{\calGapIdHalu}{+0.254}
\newcommand{\calEceTsHalu}{0.060}
\newcommand{\calEceTbHalu}{0.009}
\newcommand{\calEceIsoHalu}{0.040}
\newcommand{\dAccMagGencRbb}{0.324}
\newcommand{\dAccMagRmRbb}{0.525}
\newcommand{\dAccMagGencJb}{0.136}
\newcommand{\dAccMagRmJb}{0.289}
\newcommand{\aurocRb}{0.587}
\newcommand{\aurocLoRb}{0.562}
\newcommand{\aurocHiRb}{0.611}
\newcommand{\aurocRbb}{0.609}
\newcommand{\aurocRbp}{0.602}
\newcommand{\aurocRmb}{0.487}
\newcommand{\aurocJb}{0.552}
\newcommand{\aurocHalu}{0.495}
\newcommand{\escRb}{0.933}
\newcommand{\retRb}{0.983}
\newcommand{\cascDiffRb}{-0.014}
\newcommand{\escRbb}{1.000}
\newcommand{\retRbb}{1.000}
\newcommand{\escRbp}{0.923}
\newcommand{\retRbp}{0.989}
\newcommand{\cascDiffRbp}{-0.009}
\newcommand{\escRmb}{0.954}
\newcommand{\retRmb}{0.976}
\newcommand{\cascDiffRmb}{-0.019}
\newcommand{\escJb}{0.979}
\newcommand{\retJb}{1.000}
\newcommand{\escHalu}{1.000}
\newcommand{\retHalu}{1.000}
\newcommand{\hfiveWorst}{0.000}
\newcommand{\hsixPointRmb}{0.70}
\newcommand{\flipClm}{0.0002}
\newcommand{\flipLoClm}{0.0000}
\newcommand{\flipHiClm}{0.0008}
\newcommand{\flipNClm}{3}
\newcommand{\flipLaya}{0.4395}
\newcommand{\flipGenc}{0.2188}
\newcommand{\flipRm}{0.0019}
\newcommand{\lenAllClm}{0.476}
\newcommand{\lenGsClm}{0.453}
\newcommand{\lenDeltaClm}{-0.023}
\newcommand{\lenAllGenc}{0.559}
\newcommand{\lenGsGenc}{0.342}
\newcommand{\lenDeltaGenc}{-0.217}
\newcommand{\lenAllRm}{0.481}
\newcommand{\lenGsRm}{0.082}
\newcommand{\lenDeltaRm}{-0.399}
\newcommand{\ftDiffRb}{-0.028}
\newcommand{\ftDiffRbb}{+0.035}
\newcommand{\ftDiffRbp}{+0.054}
\newcommand{\ftDiffLoRbp}{+0.032}
\newcommand{\ftDiffHiRbp}{+0.077}
\newcommand{\ftDiffRmb}{+0.031}
\newcommand{\ftDiffJb}{+0.060}
\newcommand{\htenAcc}{0.581}
\newcommand{\htenAccLo}{0.562}
\newcommand{\htenAccHi}{0.599}
\newcommand{\htenDiff}{-0.147}
\newcommand{\cascAccRb}{0.854}
\newcommand{\rndAccRb}{0.848}
\newcommand{\cascAccRbb}{0.732}
\newcommand{\rndAccRbb}{0.732}
\newcommand{\cascAccRbp}{0.818}
\newcommand{\rndAccRbp}{0.809}
\newcommand{\cascAccRmb}{0.741}
\newcommand{\rndAccRmb}{0.749}
\newcommand{\cascAccJb}{0.615}
\newcommand{\rndAccJb}{0.613}
\newcommand{\cascAccHalu}{0.745}
\newcommand{\rndAccHalu}{0.745}
\newcommand{\agreeHalu}{0.662}
\newcommand{\accGensItemHalu}{0.745}
\newcommand{\agreeJb}{0.482}
\newcommand{\accGensItemJb}{0.615}
\newcommand{\agreeRb}{0.532}
\newcommand{\accGensItemRb}{0.868}
\newcommand{\agreeRbb}{0.339}
\newcommand{\accGensItemRbb}{0.732}
\newcommand{\agreeRbp}{0.580}
\newcommand{\accGensItemRbp}{0.827}
\newcommand{\agreeRmb}{0.475}
\newcommand{\accGensItemRmb}{0.760}

\newcommand{\clmfull}{\texttt{Contrastive-LM/CLM-v0.1-8B}}
\newcommand{\genc}{Qwen3-8B (generative)}

\newcommand{\rmj}{Skywork-Reward-V2-8B}

\title{CLM-as-a-Judge: Evaluating an Open Contrastive Decision Model on
Public Judge Benchmarks}
\hypersetup{pdftitle={CLM-as-a-Judge: Evaluating an Open Contrastive
Decision Model on Public Judge Benchmarks}}
\author{Gowthamkumar Nandakishore\\[2pt]\normalsize Independent researcher}
\date{October 2026}

\begin{document}
\maketitle

\begin{abstract}
\noindent An open contrastive decision model is near chance as a judge on
the hard public benchmarks: \clmfull{} scores between \accClmRbb{}
(best-of-four, chance \accTrandRbb{}) and \accClmRbp{} (pairwise, chance
\accTrandRbp{}), is statistically indistinguishable from coin flipping on
RM-Bench and JudgeBench, and answers every HaluEval item with one constant
label, matching the trivial always-first baseline at \accClmHalu{}. Judges
with the same parameter count score far higher everywhere: a reward model
reaches \accRmJb{} to \accRmRb{} and a generative judge \accGencJb{} to
\accGencRb{}, and every gap to CLM is significant after Benjamini--Hochberg
correction. Two properties do work. Raw confidences are overconfident by up
to \calGapIdRbb{}, yet one pooled temperature fit on held-out calibration
items repairs expected calibration error to at most \calEceTsJb{}, and the
repaired confidence ranks the model's own errors above chance on three of
six benchmarks. The decision order-flip rate is \flipClm{} against
\flipGenc{} for the generative judge, and the length-preference shift is
\lenDeltaClm{} against \lenDeltaGenc{}. The confidence-gated cascade,
however, escalates between
\escRbp{} and \escRbb{} of items to the strong judge at the preregistered
\cRetained{} retention bar: calibrated confidence about a near-chance judge
has almost nothing to keep. The design: five public preference benchmarks and
one hallucination benchmark with real labels, scored under a
preregistration frozen before any test item was seen, against generative,
reward-model, and trivial baselines, with per-item predictions released.
The author has no affiliation with, funding from, or financial interest in
Contrastive-LM, Alibaba/Qwen, Amazon/AWS, Convai Innovations, TypeSafe,
meraGPT, or the publishers of any benchmark or baseline model used.
\end{abstract}

\section{Introduction}\label{sec:intro}

Decision-only judges promise a fixed small cost per decision, a confidence
score that is a model quantity rather than a sampled token, and immunity to
the order in which candidate answers are presented. Whether an open
contrastive decision model also delivers judging competence is an empirical
question. This study answers it for \clmfull{} under a preregistration
frozen before any evaluation item was scored, against generative,
reward-model, and trivial baselines measured on the same items under the
same scoring rules.

The first finding is negative: zero-shot CLM is not a usable judge on any
benchmark studied here. It clears its trivial baselines on RewardBench,
RewardBench~2, and the RewardBench~2 pair conversion, reaching at most
\accClmRbp{} against \accTrandRbp{} chance, and it is statistically
indistinguishable from coin flipping on RM-Bench and JudgeBench. Same-size
autoregressive judges score far higher everywhere: the reward model by
\dAccMagRmJb{} to \dAccMagRmRbb{} accuracy points, the generative judge by
\dAccMagGencJb{} to \dAccMagGencRbb{}. On hallucination recognition CLM
degenerates to a single answer for every item. Fine-tuning new decision
heads on ten thousand preference pairs moves accuracy by \ftDiffRb{} to
\ftDiffJb{} per benchmark, which changes none of these conclusions.

The second finding is that the confidence channel works even though the
judge does not. Raw similarity confidences are overconfident by
\calGapIdRb{} to \calGapIdRbb{}, yet one pooled temperature fit on held-out
calibration items repairs expected calibration error to at most
\calEceTsJb{} on every benchmark, and the repaired confidence ranks the
model's own errors above chance on three of six benchmarks. Calibrated
confidence about a near-chance judge cannot gate: no preregistered
threshold rule reaches the \cErrTarget{} selective error target at usable
coverage, and the confidence-gated cascade escalates nearly every item.

The third finding is that the structural claims survive measurement. CLM's
decisions flip under answer-order swap at a rate of \flipClm{}, every flip
an exact score tie; the generative judge flips at \flipGenc{} and a smaller
decision baseline at \flipLaya{}. CLM is nearly indifferent to answer
length, where the generative and reward judges shift substantially with
it. These properties come from the architecture, not from
accuracy, and they are the only properties this study can recommend.

\paragraph{Contributions.}
\begin{enumerate}[leftmargin=1.6em,itemsep=1pt,topsep=2pt]
\item A preregistered, artifact-complete evaluation of an open contrastive
decision model as a judge on six public benchmarks against generative,
reward-model, and trivial baselines: CLM trails the generative judge by
\dAccMagGencJb{} to \dAccMagGencRbb{} and the reward model by
\dAccMagRmJb{} to \dAccMagRmRbb{}, all gaps significant at $q=\cBhQ$.
\item Calibration and error-detection measurements for embedding-similarity
confidence: one pooled temperature (\calTsingle) repairs expected
calibration error from up to \calEceIdRbb{} to at most \calEceTsJb{}.
\item Measured order-flip and length-bias rates for four judge families:
\flipClm{} for CLM against \flipGenc{} for the generative judge.
\item A confidence-gated cascade reported as agreement with the strong
judge, escalation fraction, and retained accuracy: the gate escalates
between \escRbp{} and \escRbb{} of items at the \cRetained{} retention bar.
\item Released per-item predictions, manifests, metric code, and a frozen
preregistration; every printed number maps mechanically to an artifact key.
\end{enumerate}

The author has no affiliation with, funding from, or financial interest in
Contrastive-LM, Alibaba/Qwen, Amazon/AWS, Convai Innovations, TypeSafe,
meraGPT, or the publishers of any benchmark or baseline model used.

\section{Related work}\label{sec:related}

JEV-as-a-Judge~\citep{li2026jev} evaluates a small decision model as a
judge on public benchmarks and escalates to a stronger judge below a
confidence threshold; this study follows that design (public benchmarks
with real labels, generative and reward-model comparators, calibration and
error detection, a measured cascade) and applies it to a different model
class. No number from that paper is compared against a measured number
here: its judges, hardware, and prompts differ. LLM-as-a-Verifier
\citep{kwok2026verifier} studies verification as a general interface;
BERT-as-a-Judge~\citep{gisserot2026bert} shows a small encoder can replace
lexical metrics for reference-based evaluation. CLM is described in its
technical report~\citep{kwok2026clm}; a companion study
\citep{nandakishore2026laya} established the reproduction protocol reused
here.

The benchmarks are RewardBench~\citep{lambert2024rewardbench},
RewardBench~2~\citep{malik2025rewardbench2}, RM-Bench~\citep{liu2025rmbench},
JudgeBench~\citep{tan2025judgebench}, and HaluEval~\citep{li2023halueval};
each benchmark's official scoring rule is ground truth. Generative judges
show position and length bias~\citep{zheng2023mtbench,wang2024fair} and
favor their own generations~\citep{panickssery2024own}; agreement with a
stronger teacher does not establish correctness~\citep{gudibande2023imitation}.
Judgment detection work shows LLM-generated judgments are identifiable from
score-candidate interactions alone, which quantifies judge
bias~\citep{li2025judgedetect}. The architecture studied here scores
candidates independently of presentation order, making position bias a
testable structural claim. Temperature scaling and its evaluation are
standard~\citep{guo2017calibration,naeini2015bayesian,nixon2019measuring,
roelofs2022mitigating,zadrozny2002isotonic}; selective prediction and
learning to defer formalize accept-or-escalate behavior
\citep{chow1970reject,elyaniv2010selective,geifman2017selective,
mozannar2020defer}, and LLM cascades apply it to model pairs
\citep{chen2024frugalgpt,ong2024routellm,jitkrittum2023cascade,
li2021cascadebert,dekoninck2024unified}.

\section{Setup}\label{sec:setup}

All models and datasets are public. Every locally run model is pinned by
immutable revision hash in \texttt{anc/pins.json}; local judges run on one
Apple M5 Max with 64\,GB of unified memory, and the strong judge is hosted
(below). Latency is reported for local judges only, as context; the hosted
judge's latency is service latency and no cost comparison against it is
made.

\paragraph{CLM.}
\clmfull{} selects among typed candidate answers by embedding similarity
behind a local embeddings server written for this study; the encoder is not
quantized, and its serving path was verified against the publisher's
released examples before any scoring (one discrepancy with the publisher's
quickstart prints is recorded in \texttt{anc/metrics/}). Judgments are zero
shot: a pairwise item becomes one typed choice question with two
candidates, a best-of-four item one choice question with four candidates,
and a hallucination item one typed binary question. CLM reports a
probability per candidate; its confidence is the maximum probability. A
fine-tuned variant (new decision heads trained on \cHeadsN{} pairs from the
pinned Skywork preference dataset~\citep{liu2024skywork}, deduplicated
against benchmark prompts, frozen encoder untouched) runs the same passes.

\paragraph{laya.}
The laya 0.3.20 base checkpoint~\citep{nandakishore2026laya} answers the
same typed questions through its own runtime, with a 512-token content
window.

\paragraph{Cheap generative judge.}
Qwen3-8B~\citep{qwen3report} in non-thinking chat mode via MLX in bfloat16
\citep{mlx2023}, greedy decoding, at most 64 new tokens, one frozen prompt
per task form requesting a one-line JSON label with a stated confidence; a
reply that does not parse falls back to the first valid label at confidence
0.5 and is flagged.

\paragraph{Strong judge.}
{\sloppy Qwen3-32B (strong judge), served through Amazon Bedrock as
\texttt{qwen.\allowbreak qwen3-32b-v1:0}, temperature 0, 64 completion
tokens, frozen prompt; access dates and token counts in \texttt{anc/}.\par}

\paragraph{Reward model.}
\rmj{}~\citep{liu2025skyworkv2}, a sequence classifier, selected by a
preregistered rule: the top open sequence-classifier reward model at or
under 8B parameters by overall mean on the pinned RewardBench~2 leaderboard
snapshot. A pairwise probability is the logistic of the score difference;
best-of-four uses a softmax over scores. A preference reward model defines
no calibrated binary judgment for a single response, so it answers no
HaluEval items (a preregistered exclusion).

\paragraph{Trivial baselines.}
Always-first, a per-item-seeded random choice, and longer-response-wins.

\paragraph{Benchmarks and task mapping.}
Five preference benchmarks and one hallucination benchmark, all with real
labels, all pinned: RewardBench (filtered split), RewardBench~2
(best-of-four primary; a derived chosen-versus-each-rejected pairwise form
secondary), RM-Bench (official style matrix; the aggregate is the mean of
the four domain averages), JudgeBench, and HaluEval. The RewardBench~2 Ties
subset is excluded: its official score is a weighted composite with no
per-item accuracy. RewardBench~2 credit follows the official rule: a tied
maximum that includes the chosen response earns the reciprocal of the tie
count. HaluEval becomes a binary typed question; its qa, dialogue, and
summarization sub-tasks are subsampled to 1{,}000 rows each with the study
seed by a deterministic per-item coin, and the general sub-task is kept as
is. A deterministic per-item coin places the labeled winner at position A
or B; pairwise and best-of-four items run in both presentation orders,
hallucination items and the strong judge in order 1 only; analyses
involving the strong judge use order-1 passes for every tier.

\paragraph{Truncation.}
One policy for every judge: the prompt is kept whole up to the smaller of
512 tokens and half the judge's content window; the remaining window is
split equally across response texts; every text keeps its tail. Windows are
\cWinClm{} tokens for CLM, \cWinLaya{} for laya, and \cWinGen{} for the
generative judges and the reward model, counted by each judge's own
tokenizer. Table~\ref{tab:trunc} reports the truncated fraction per family
and benchmark; the laya window truncates most items on four benchmarks and
CLM truncates a majority of RewardBench~2 items, which frames every
comparison that follows.

\begin{table}[t]
\centering
\caption{Truncated items per judge family and benchmark: fraction of items
with any truncated text (left of slash) and fraction of input tokens kept
(right), measured with each family's own tokenizer under the single frozen
policy. Source: \texttt{anc/metrics/gate05\_truncation.json}.}
\label{tab:trunc}
\small
\begin{tabular}{lcccc}
\toprule
Benchmark & CLM (\cWinClm) & laya (\cWinLaya) & Generative (\cWinGen) & RM (\cWinRm) \\
\midrule
RewardBench & \truncClmRb{}\,/\,\keptClmRb & \truncLayaRb{}\,/\,\keptLayaRb & \truncGenRb{}\,/\,\keptGenRb & \truncRmRb{}\,/\,\keptRmRb \\
RewardBench~2 & \truncClmRbb{}\,/\,\keptClmRbb & \truncLayaRbb{}\,/\,\keptLayaRbb & \truncGenRbb{}\,/\,\keptGenRbb & \truncRmRbb{}\,/\,\keptRmRbb \\
RewardBench~2 pairs & \truncClmRbp{}\,/\,\keptClmRbp & \truncLayaRbp{}\,/\,\keptLayaRbp & \truncGenRbp{}\,/\,\keptGenRbp & \truncRmRbp{}\,/\,\keptRmRbp \\
RM-Bench & \truncClmRmb{}\,/\,\keptClmRmb & \truncLayaRmb{}\,/\,\keptLayaRmb & \truncGenRmb{}\,/\,\keptGenRmb & \truncRmRmb{}\,/\,\keptRmRmb \\
JudgeBench & \truncClmJb{}\,/\,\keptClmJb & \truncLayaJb{}\,/\,\keptLayaJb & \truncGenJb{}\,/\,\keptGenJb & \truncRmJb{}\,/\,\keptRmJb \\
HaluEval & \truncClmHalu{}\,/\,\keptClmHalu & \truncLayaHalu{}\,/\,\keptLayaHalu & \truncGenHalu{}\,/\,\keptGenHalu & \truncRmHalu{}\,/\,\keptRmHalu \\
\bottomrule
\end{tabular}
\end{table}

\paragraph{Metrics and statistics.}
Accuracy is mean per-item credit under each benchmark's official rule.
Calibration uses NLL, equal-width 15-bin ECE, and the signed gap (mean
confidence minus accuracy)
\citep{guo2017calibration,nixon2019measuring,roelofs2022mitigating,gneiting2007scoring};
error detection uses the AUROC of confidence against per-item correctness.
Every interval is a cluster bootstrap at prompt level with \cBootN{}
resamples and seed 0: BCa for single quantities, percentile for paired
deltas, add-one two-sided
p-values~\citep{efron1993bootstrap,davison1997bootstrap}. The preregistered
hypothesis family is tested with Benjamini--Hochberg control at
$q=\cBhQ$~\citep{benjamini1995bh}. Items sharing a prompt always resample
together.

\paragraph{Preregistration.}
The preregistration was frozen before any evaluation item was scored; its
SHA-256 and freeze time are in \texttt{anc/prereg\_freeze.json}, and every
evaluation manifest postdates the freeze~\citep{nosek2018prereg}. Splits
are 30\% calibration and 70\% evaluation, stratified by benchmark and
subset, assigned at prompt level with seed \cSeed{}; the split digest and
per-benchmark listings are released. Temperatures, isotonic maps, and every
threshold are fit on calibration items only. Two departures from the
preregistration occurred. The strong judge ran on Amazon Bedrock rather
than locally, so the preregistered latency-cost criterion for H4 was not
evaluated. Duplicate source ids in two benchmarks were handled by
occurrence index, so twin items are scored and aligned exactly once each.

\section{Results: accuracy}\label{sec:results-acc}

Table~\ref{tab:acc} reports evaluation-split accuracy for every judge on
every benchmark under the official scoring rules, with BCa 95\% intervals
for all cells in \texttt{anc/metrics/comparator\_table.json}. CLM clears
its trivial baselines on three benchmarks: \accClmRb{} on RewardBench
(interval \accLoClmRb{} to \accHiClmRb{}), \accClmRbb{} on RewardBench~2
best-of-four (chance \accTrandRbb{}), and \accClmRbp{} on the pair
conversion. On RM-Bench (\accClmRmb{}, interval \accLoClmRmb{} to
\accHiClmRmb{}) and JudgeBench (\accClmJb{}, interval \accLoClmJb{} to
\accHiClmJb{}) both intervals contain chance: on the two benchmarks built
around subtle content differences and objective correctness the
embedding-similarity judge carries no signal this design can detect. Every
same-size autoregressive judge is far ahead: all ten preregistered H1
differences are significant, and CLM trails the generative judge by
\dAccMagGencJb{} to \dAccMagGencRbb{} and the reward model by
\dAccMagRmJb{} to \dAccMagRmRbb{}. The hosted strong judge scores between
\accGensJb{} and
\accGensRb{}. The laya baseline sits at or near chance everywhere, never
more than a few points above it (Table~\ref{tab:acc}), consistent with its
window truncating most inputs (Table~\ref{tab:trunc});
the length baseline is below random everywhere, so gold responses here are
more often the shorter ones. Judges not named in a preregistered hypothesis
are compared descriptively only.

\begin{table}[t]
\centering
\caption{Evaluation accuracy per judge and benchmark (measured). Chance is
\accTrandRbb{} on RewardBench~2 best-of-four and \accTrandRb{} on the
pairwise benchmarks. The strong judge is hosted; all other judges ran
locally. Source: \texttt{anc/metrics/comparator\_table.json}, which also
carries the BCa 95\% intervals.}
\label{tab:acc}
\footnotesize
\setlength{\tabcolsep}{3.5pt}
\begin{tabular}{lcccccc}
\toprule
Judge & RewardBench & RewardBench~2 & RB2 pairs & RM-Bench & JudgeBench & HaluEval \\
\midrule
CLM & \accClmRb & \accClmRbb & \accClmRbp & \accClmRmb & \accClmJb & \accClmHalu \\
CLM (tuned heads) & \accClmftRb & \accClmftRbb & \accClmftRbp & \accClmftRmb & \accClmftJb & \accClmftHalu \\
laya & \accLayaRb & \accLayaRbb & \accLayaRbp & \accLayaRmb & \accLayaJb & \accLayaHalu \\
\genc{} & \accGencRb & \accGencRbb & \accGencRbp & \accGencRmb & \accGencJb & \accGencHalu \\
Qwen3-32B (strong, hosted) & \accGensRb & \accGensRbb & \accGensRbp & \accGensRmb & \accGensJb & \accGensHalu \\
\rmj{} & \accRmRb & \accRmRbb & \accRmRbp & \accRmRmb & \accRmJb & --- \\
Trivial: always first & \accTfirstRb & \accTfirstRbb & \accTfirstRbp & \accTfirstRmb & \accTfirstJb & \accTfirstHalu \\
Trivial: random & \accTrandRb & \accTrandRbb & \accTrandRbp & \accTrandRmb & \accTrandJb & \accTrandHalu \\
Trivial: length & \accTlenRb & \accTlenRbb & \accTlenRbp & \accTlenRmb & \accTlenJb & \accTlenHalu \\
\bottomrule
\end{tabular}
\end{table}

\paragraph{Fine-tuned heads.}
A day of preference data moves the judge, but not out of its class. The
tuned heads reach validation accuracy \headsValAcc{} and holdout accuracy
\headsTestAcc{} against \headsInitTest{} for the released heads on the
training distribution, and on the benchmarks they change accuracy by
\ftDiffRbp{} on the pair conversion (interval \ftDiffLoRbp{} to
\ftDiffHiRbp{}), \ftDiffJb{} on JudgeBench, \ftDiffRmb{} on RM-Bench,
\ftDiffRbb{} on RewardBench~2, and \ftDiffRb{} on RewardBench, where the
tuned heads are worse than the released ones. All five differences are
significant (family H9), so the training signal is real in both directions;
the largest gain is several times smaller than the gaps to the
autoregressive judges.

\paragraph{Hallucination detection.}
CLM answers every HaluEval item with the same label. Its accuracy of
\htenAcc{} (interval \htenAccLo{} to \htenAccHi{}) is exactly the trivial
always-first baseline (\accTfirstHalu{}): per-class accuracy in the
released predictions is one on items whose gold label matches the constant
answer and zero on the rest. The preregistered H10 test against the
sampling-coin chance of one half is significant, but the comparison that
matters is the constant baseline, and there CLM adds nothing. The
generative judge reaches \accGencHalu{} (\htenDiff{} for CLM against it,
significant) and the strong judge \accGensHalu{}. The tuned heads produce
the same constant-label behavior.

\paragraph{Order and length stability.}
Each pairwise and best-of-four item was scored in both presentation orders
in separate passes. CLM's flip rate is \flipClm{} (BCa interval
\flipLoClm{} to \flipHiClm{}; \flipNClm{} flips, each an exact score tie
broken by position). The reward model, which also scores responses one at a
time, flips at \flipRm{}; the generative judge flips at \flipGenc{}, more
than one in five items; laya flips at \flipLaya{}. All three differences
against CLM are significant. On length: over all pair items CLM chooses the
longer response at \lenAllClm{}, falling only to \lenGsClm{} where the gold
response is the shorter one (delta \lenDeltaClm{}), so CLM is indifferent
to length and nearly indifferent to correctness. The generative judge moves
from \lenAllGenc{} to \lenGsGenc{} (delta \lenDeltaGenc{}) and the reward
model from \lenAllRm{} to \lenGsRm{} (delta \lenDeltaRm{}): both track the
gold answer away from the longer response.

\section{Results: calibration, error detection, and gating}\label{sec:results-cal}

\begin{figure}[t]
\centering
\includegraphics[width=\linewidth]{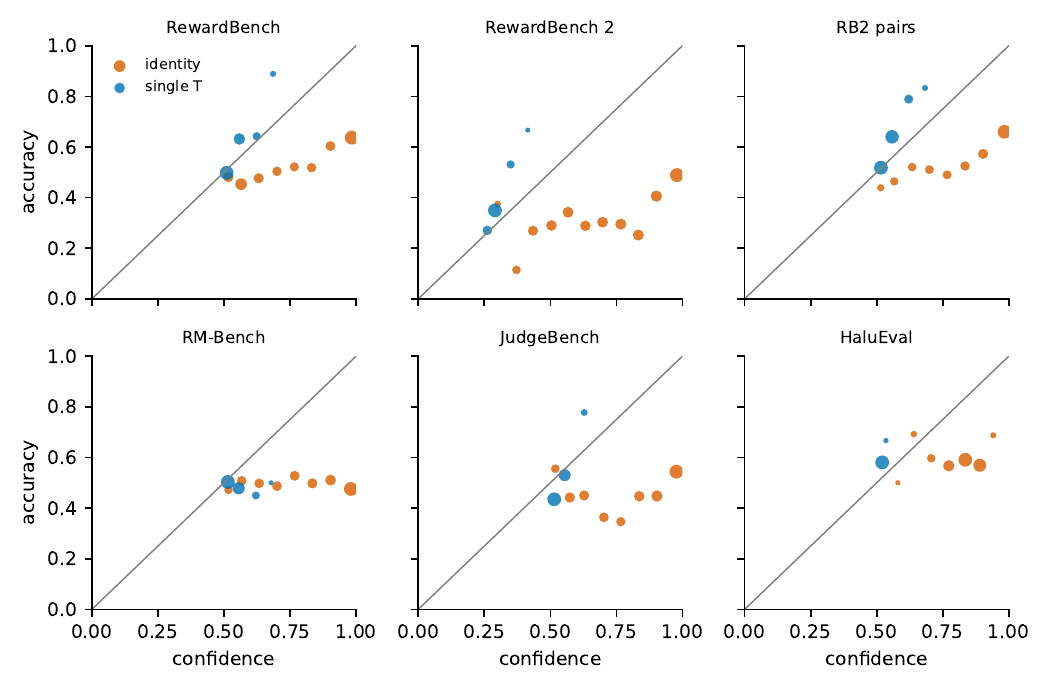}
\caption{Reliability diagrams for CLM confidence on evaluation items,
identity map against the pooled temperature map, per benchmark. Generated
by \texttt{anc/code/build\_figures.py} from released predictions.}
\label{fig:reliability}
\end{figure}

Four confidence maps are compared, all fit on calibration items only:
identity, one pooled temperature, one temperature per benchmark, and
per-benchmark isotonic regression
\citep{guo2017calibration,zadrozny2002isotonic}. Temperature preserves the
argmax, so accuracy is unchanged; the question is whether confidence
becomes usable. Raw confidence is severely overconfident everywhere: the
signed gap runs from \calGapIdRb{} on RewardBench to \calGapIdRbb{} on
RewardBench~2, and all eighteen preregistered H2 contrasts are significant.
The pooled temperature is reported as fitted: \calTsingle{}, which sits at
the upper bound of the fitter's search range ($\log T \in [-3, 3]$, so $T
\le e^{3}$); the unconstrained optimum would be somewhat larger, every
reported map uses the fitted value consistently, and no map was refit for
this paper. One pooled temperature is enough: expected calibration error
falls from \calEceIdRb{}--\calEceIdRbb{} under identity to between
\calEceTsRb{} and \calEceTsJb{} (Table~\ref{tab:cal},
Figure~\ref{fig:reliability}); per-benchmark temperatures and isotonic maps
help on some benchmarks and hurt on JudgeBench.

\begin{table}[t]
\centering
\caption{Expected calibration error on evaluation items under the four
confidence maps, all fit on calibration items only. The identity column
also shows the signed confidence-minus-accuracy gap. Source:
\texttt{anc/metrics/calibration\_table.json}.}
\label{tab:cal}
\small
\begin{tabular}{lccccc}
\toprule
Benchmark & Identity ECE & Signed gap & Pooled $T$ & Per-bench $T$ & Isotonic \\
\midrule
RewardBench & \calEceIdRb & \calGapIdRb & \calEceTsRb & \calEceTbRb & \calEceIsoRb \\
RewardBench~2 & \calEceIdRbb & \calGapIdRbb & \calEceTsRbb & \calEceTbRbb & \calEceIsoRbb \\
RewardBench~2 pairs & \calEceIdRbp & \calGapIdRbp & \calEceTsRbp & \calEceTbRbp & \calEceIsoRbp \\
RM-Bench & \calEceIdRmb & \calGapIdRmb & \calEceTsRmb & \calEceTbRmb & \calEceIsoRmb \\
JudgeBench & \calEceIdJb & \calGapIdJb & \calEceTsJb & \calEceTbJb & \calEceIsoJb \\
HaluEval & \calEceIdHalu & \calGapIdHalu & \calEceTsHalu & \calEceTbHalu & \calEceIsoHalu \\
\bottomrule
\end{tabular}
\end{table}

Calibrated confidence ranks the model's own errors above chance (AUROC
against a permutation null) on RewardBench (\aurocRb{}, interval
\aurocLoRb{} to \aurocHiRb{}), RewardBench~2 (\aurocRbb{}), and the pair
conversion (\aurocRbp{}); the JudgeBench point estimate (\aurocJb{}) does
not survive correction, and RM-Bench (\aurocRmb{}) and HaluEval
(\aurocHalu{}) sit at chance. Where the judge is at chance, its confidence
has no errors to rank. Gating is unambiguous and negative: at the frozen
error target of \cErrTarget{}, none of the three preregistered acceptance
rules (point, cross-fit, Wilson upper bound~\citep{wilson1927probable})
produces a threshold on five of six benchmarks, because with accuracy near
chance even the most confident slice carries error far above the target;
RM-Bench alone yields \hsixPointRmb{} under the point and cross-fit rules.

\section{Cascade}\label{sec:cascade}

The cascade answers each item with CLM and escalates to the strong judge
when CLM's calibrated confidence falls below a threshold; the threshold is
the lowest calibration-split grid point whose retained accuracy clears the
preregistered \cRetained{} bar. Because the strong judge is hosted, the
preregistered latency-cost criterion for H4 was not evaluated; what is
reported is hardware-independent and framed as agreement with the strong
judge. Table~\ref{tab:cascade} gives, per benchmark: CLM-alone accuracy,
strong-judge-alone accuracy, the agreement rate between CLM and the strong
judge on evaluation items, cascade accuracy at the retention-selected
threshold, the escalation fraction, the retained fraction, and the
random-escalation control at the matched fraction.

\begin{table}[t]
\centering
\caption{Cascade as agreement with the strong judge, per benchmark:
accuracy of each tier alone, rate at which both judges emit the same
decision, cascade accuracy at the retention-selected threshold $\tau^{*}$,
escalation fraction, retained fraction of strong-judge accuracy, and the
random-escalation control at the matched fraction. Strong, Cascade,
Random, and Retained all use one convention, the item-level mean on
order-1 evaluation items; RM-Bench's value in Table~\ref{tab:acc} instead
uses the benchmark's official domain-average aggregate. Sources:
\texttt{anc/metrics/comparator\_table.json},
\texttt{cascade\_agreement.json}, \texttt{h4\_cascade.json}.}
\label{tab:cascade}
\small
\setlength{\tabcolsep}{4.5pt}
\begin{tabular}{lccccccc}
\toprule
Benchmark & CLM & Strong & Agree & Cascade & Escal. & Retained & Random \\
\midrule
RewardBench & \accClmRb & \accGensItemRb & \agreeRb & \cascAccRb & \escRb & \retRb & \rndAccRb \\
RewardBench~2 & \accClmRbb & \accGensItemRbb & \agreeRbb & \cascAccRbb & \escRbb & \retRbb & \rndAccRbb \\
RewardBench~2 pairs & \accClmRbp & \accGensItemRbp & \agreeRbp & \cascAccRbp & \escRbp & \retRbp & \rndAccRbp \\
RM-Bench & \accClmRmb & \accGensItemRmb & \agreeRmb & \cascAccRmb & \escRmb & \retRmb & \rndAccRmb \\
JudgeBench & \accClmJb & \accGensItemJb & \agreeJb & \cascAccJb & \escJb & \retJb & \rndAccJb \\
HaluEval & \accClmHalu & \accGensItemHalu & \agreeHalu & \cascAccHalu & \escHalu & \retHalu & \rndAccHalu \\
\bottomrule
\end{tabular}
\end{table}

\begin{figure}[t]
\centering
\includegraphics[width=\linewidth]{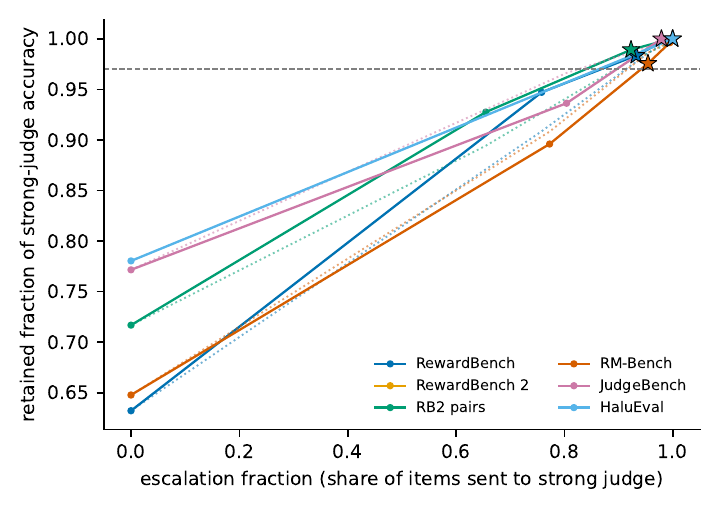}
\caption{Retained accuracy against escalation fraction for the
confidence-gated cascade (per grid threshold) and the matched
random-escalation control, per benchmark. The horizontal line is the
preregistered \cRetained{} retention bar. Generated by
\texttt{anc/code/build\_figures.py}.}
\label{fig:cascade}
\end{figure}

The gate is nearly decorative (Figure~\ref{fig:cascade}). CLM agrees with
the strong judge on \agreeRbb{} to \agreeHalu{} of items, roughly its own
accuracy level, so low confidence is the norm and the retention-selected
threshold escalates \escRbp{} to \escRbb{} of items; on RewardBench~2,
JudgeBench, and HaluEval the cascade is the strong judge. Where the gate
keeps items it gives up accuracy against the strong judge alone
(\cascDiffRb{} on RewardBench, \cascDiffRbp{} on the pair conversion,
\cascDiffRmb{} on RM-Bench, each significant), and the random-escalation
control at the matched fraction performs comparably, so the gate's
targeting adds little beyond its volume. Threshold transfer collapses for
the same reason: only RM-Bench produced a threshold, and applied elsewhere
it accepts almost nothing (at most six items on any target benchmark,
accepted error \hfiveWorst{}). Table~\ref{tab:latency} reports local
judges' latencies as context; no cost ratio against the hosted strong judge
is computed.

\begin{table}[t]
\centering
\caption{Measured median latency per item in milliseconds for the local
judges (20-item probes, first two warm-up items skipped, one Apple M5 Max),
shown as context only. The artifact stores medians at one-tenth-millisecond
resolution; printed integers round the stored values. Source:
\texttt{anc/metrics/gate03\_latency.json}.}
\label{tab:latency}
\small
\begin{tabular}{lcccc}
\toprule
Benchmark & laya & CLM & \rmj{} & \genc{} \\
\midrule
RewardBench & \latLayaRb & \latClmRb & \latRmRb & \latGencRb \\
RewardBench~2 & \latLayaRbb & \latClmRbb & \latRmRbb & \latGencRbb \\
RewardBench~2 pairs & \latLayaRbp & \latClmRbp & \latRmRbp & \latGencRbp \\
RM-Bench & \latLayaRmb & \latClmRmb & \latRmRmb & \latGencRmb \\
JudgeBench & \latLayaJb & \latClmJb & \latRmJb & \latGencJb \\
HaluEval & \latLayaHalu & \latClmHalu & --- & \latGencHalu \\
\bottomrule
\end{tabular}
\end{table}

\section{Limitations}\label{sec:limitations}

\begin{enumerate}[leftmargin=1.6em,itemsep=1pt,topsep=2pt]
\item \textbf{Hosted strong judge.} The strong judge is served by Amazon
Bedrock; provider weights, precision, and serving configuration are not
revision-pinned and can change without notice, so its results are a
property of the service at the recorded access dates.
\item \textbf{Encoder serving path.} The CLM encoder runs on a locally
built serving path verified against the publisher's released examples; one
discrepancy with the publisher's quickstart prints is recorded in
\texttt{anc/metrics/}. If the reference path differs in some unprobed way,
CLM's numbers could shift.
\item \textbf{Truncation.} The 512-token window truncates most laya inputs
on four of six benchmarks; CLM truncates a majority of RewardBench~2
inputs. Rankings between judges with different windows partly reflect
window size.
\item \textbf{No human adjudication.} Benchmark labels are ground truth,
including their known noise.
\item \textbf{Scope.} One cheap judge, one strong judge, one error target;
local latencies are specific to one machine and these serving stacks;
hosted frontier judges beyond the one used were not run, so nothing here
bounds what they would do.
\end{enumerate}

\section{Ethics}\label{sec:ethics}

This study evaluates released models on released benchmarks with real
labels; no human subjects, private data, or crowdwork were involved.
Benchmark contents include adversarial and safety-sensitive text by design;
nothing new of that kind was generated. Energy use was one workstation for
the study's duration plus one hosted judge.

\section{Independence statement}\label{sec:independence}
The author has no affiliation with, funding from, or financial interest in
Contrastive-LM, Alibaba/Qwen, Amazon/AWS, Convai Innovations, TypeSafe,
meraGPT, or the publishers of any benchmark or baseline model used.

\section{Reproducibility}\label{sec:repro}

The ancillary files (\texttt{anc/}) contain the preregistration and its
freeze record, all pins with revision hashes, split listings and digests,
one manifest per pass, per-item predictions for every judge and benchmark,
all metric JSONs, and the metric and analysis code. Every number in every
table and in the abstract is a generated macro mapping to a named ancillary
file and key; the mapping was checked mechanically
(\texttt{anc/NUMBER\_CHECK.md}) and the reported quantities were recomputed
from the artifacts alone by independent sessions with no access to the
manuscript source. Local models and datasets load by pinned revision; the
study seed is \cSeed{}; the bootstrap seed is 0. Reproducing the analysis
from the released predictions requires no credential. Re-scoring the strong
judge's passes requires an Amazon Bedrock credential and reaches a service
that is not revision-pinned, so bit-exact reproduction of those predictions
is not guaranteed; model id, region, decoding parameters, access dates, and
token usage are in the manifests and \texttt{anc/pins.json}.

\section*{Acknowledgements and AI-assistance disclosure}
The experiments were executed by LLM agents under the author's direction.
Every reported number was verified against the released artifacts by three
independent passes (Section~\ref{sec:repro}). The author reviewed the
manuscript and takes responsibility for its content.

\bibliographystyle{plainnat}
\bibliography{refs}

\begin{thebibliography}{39}
\providecommand{\natexlab}[1]{#1}
\providecommand{\url}[1]{\texttt{#1}}
\expandafter\ifx\csname urlstyle\endcsname\relax
  \providecommand{\doi}[1]{doi: #1}\else
  \providecommand{\doi}{doi: \begingroup \urlstyle{rm}\Url}\fi

\bibitem[Benjamini and Hochberg(1995)]{benjamini1995bh}
Yoav Benjamini and Yosef Hochberg.
\newblock Controlling the false discovery rate: A practical and powerful approach to multiple testing.
\newblock \emph{Journal of the Royal Statistical Society: Series B}, 57\penalty0 (1):\penalty0 289--300, 1995.

\bibitem[Chen et~al.(2024)Chen, Zaharia, and Zou]{chen2024frugalgpt}
Lingjiao Chen, Matei Zaharia, and James Zou.
\newblock Frugalgpt: How to use large language models while reducing cost and improving performance.
\newblock \emph{Transactions on Machine Learning Research}, 2024.
\newblock arXiv:2305.05176.

\bibitem[Chow(1970)]{chow1970reject}
C.~K. Chow.
\newblock On optimum recognition error and reject tradeoff.
\newblock \emph{IEEE Transactions on Information Theory}, 16\penalty0 (1):\penalty0 41--46, 1970.

\bibitem[Davison and Hinkley(1997)]{davison1997bootstrap}
Anthony~C. Davison and David~V. Hinkley.
\newblock \emph{Bootstrap Methods and their Application}.
\newblock Cambridge University Press, 1997.

\bibitem[Dekoninck et~al.(2024)Dekoninck, Baader, and Vechev]{dekoninck2024unified}
Jasper Dekoninck, Maximilian Baader, and Martin Vechev.
\newblock A unified approach to routing and cascading for {LLM}s, 2024.

\bibitem[Efron and Tibshirani(1993)]{efron1993bootstrap}
Bradley Efron and Robert~J. Tibshirani.
\newblock \emph{An Introduction to the Bootstrap}.
\newblock Chapman \& Hall, 1993.

\bibitem[El-Yaniv and Wiener(2010)]{elyaniv2010selective}
Ran El-Yaniv and Yair Wiener.
\newblock On the foundations of noise-free selective classification.
\newblock \emph{Journal of Machine Learning Research}, 11:\penalty0 1605--1641, 2010.

\bibitem[Geifman and El-Yaniv(2017)]{geifman2017selective}
Yonatan Geifman and Ran El-Yaniv.
\newblock Selective classification for deep neural networks.
\newblock In \emph{Advances in Neural Information Processing Systems (NeurIPS)}, 2017.

\bibitem[Gisserot-Boukhlef et~al.(2026)Gisserot-Boukhlef, Boizard, Malherbe, Hudelot, and Colombo]{gisserot2026bert}
Hippolyte Gisserot-Boukhlef, Nicolas Boizard, Emmanuel Malherbe, C{\'e}line Hudelot, and Pierre Colombo.
\newblock {BERT}-as-a-judge: A robust alternative to lexical methods for efficient reference-based {LLM} evaluation.
\newblock In \emph{Proceedings of the Conference on Language Modeling (COLM)}, 2026.
\newblock Camera-ready (arXiv:2604.09497v2, 20 July 2026).

\bibitem[Gneiting and Raftery(2007)]{gneiting2007scoring}
Tilmann Gneiting and Adrian~E. Raftery.
\newblock Strictly proper scoring rules, prediction, and estimation.
\newblock \emph{Journal of the American Statistical Association}, 102\penalty0 (477):\penalty0 359--378, 2007.

\bibitem[Gudibande et~al.(2023)Gudibande, Wallace, Snell, Geng, Liu, Abbeel, Levine, and Song]{gudibande2023imitation}
Arnav Gudibande, Eric Wallace, Charlie Snell, Xinyang Geng, Hao Liu, Pieter Abbeel, Sergey Levine, and Dawn Song.
\newblock The false promise of imitating proprietary {LLM}s, 2023.

\bibitem[Guo et~al.(2017)Guo, Pleiss, Sun, and Weinberger]{guo2017calibration}
Chuan Guo, Geoff Pleiss, Yu~Sun, and Kilian~Q. Weinberger.
\newblock On calibration of modern neural networks.
\newblock In \emph{International Conference on Machine Learning (ICML)}, 2017.
\newblock arXiv:1706.04599.

\bibitem[Hannun et~al.(2023)Hannun, Digani, Katharopoulos, and Collobert]{mlx2023}
Awni Hannun, Jagrit Digani, Angelos Katharopoulos, and Ronan Collobert.
\newblock {MLX}: Efficient and flexible machine learning on apple silicon.
\newblock \url{https://github.com/ml-explore/mlx}, 2023.
\newblock Version pinned in the ancillary files. Accessed 2026-09-28.

\bibitem[Jitkrittum et~al.(2023)Jitkrittum, Gupta, Menon, Narasimhan, Rawat, and Kumar]{jitkrittum2023cascade}
Wittawat Jitkrittum, Neha Gupta, Aditya~Krishna Menon, Harikrishna Narasimhan, Ankit~Singh Rawat, and Sanjiv Kumar.
\newblock When does confidence-based cascade deferral suffice?
\newblock In \emph{Advances in Neural Information Processing Systems (NeurIPS)}, 2023.

\bibitem[Kwok et~al.(2026{\natexlab{a}})Kwok, Kang, Suresh, Saad-Falcon, Pavone, R{\'e}, and Mirhoseini]{kwok2026clm}
Jacky Kwok, Hangoo Kang, Tarun Suresh, Jon Saad-Falcon, Marco Pavone, Christopher R{\'e}, and Azalia Mirhoseini.
\newblock Contrastive language models: A system one model for fast and generalizable decision-making.
\newblock Technical report, 2026{\natexlab{a}}.
\newblock URL \url{https://contrastive-lm.notion.site}.
\newblock Model weights: \texttt{Contrastive-LM/CLM-v0.1-8B} on Hugging Face, Apache 2.0. Accessed 2026-09-28.

\bibitem[Kwok et~al.(2026{\natexlab{b}})Kwok, Li, Atreya, Liu, Jiang, Finn, Pavone, Stoica, and Mirhoseini]{kwok2026verifier}
Jacky Kwok, Shulu Li, Pranav Atreya, Yuejiang Liu, Yixing Jiang, Chelsea Finn, Marco Pavone, Ion Stoica, and Azalia Mirhoseini.
\newblock {LLM}-as-a-verifier: A general-purpose verification framework, 2026{\natexlab{b}}.
\newblock v2, 7 Jul 2026.

\bibitem[Lambert et~al.(2024)Lambert, Pyatkin, Morrison, Miranda, Lin, Chandu, Dziri, Kumar, Zick, Choi, Smith, and Hajishirzi]{lambert2024rewardbench}
Nathan Lambert, Valentina Pyatkin, Jacob Morrison, LJ~Miranda, Bill~Yuchen Lin, Khyathi Chandu, Nouha Dziri, Sachin Kumar, Tom Zick, Yejin Choi, Noah~A. Smith, and Hannaneh Hajishirzi.
\newblock Rewardbench: Evaluating reward models for language modeling, 2024.

\bibitem[Li et~al.(2025)Li, Tan, Zhao, Jiang, Huang, Ma, Alnaibari, Shu, and Liu]{li2025judgedetect}
Dawei Li, Zhen Tan, Chengshuai Zhao, Bohan Jiang, Baixiang Huang, Pingchuan Ma, Abdullah Alnaibari, Kai Shu, and Huan Liu.
\newblock Who's your judge? on the detectability of {LLM}-generated judgments, 2025.
\newblock arXiv:2509.25154, September 2025.

\bibitem[Li et~al.(2023)Li, Cheng, Zhao, Nie, and Wen]{li2023halueval}
Junyi Li, Xiaoxue Cheng, Wayne~Xin Zhao, Jian-Yun Nie, and Ji-Rong Wen.
\newblock Halueval: A large-scale hallucination evaluation benchmark for large language models.
\newblock In \emph{Proceedings of the 2023 Conference on Empirical Methods in Natural Language Processing (EMNLP)}, pages 6449--6464, 2023.
\newblock arXiv:2305.11747.

\bibitem[Li et~al.(2021)Li, Lin, Chen, Ren, Li, Zhou, and Sun]{li2021cascadebert}
Lei Li, Yankai Lin, Deli Chen, Shuhuai Ren, Peng Li, Jie Zhou, and Xu~Sun.
\newblock Cascadebert: Accelerating inference of pre-trained language models via calibrated complete models cascade.
\newblock In \emph{Findings of the Association for Computational Linguistics: EMNLP}, 2021.

\bibitem[Li et~al.(2026)Li, Miao, Krishnan, and Padman]{li2026jev}
Yubo Li, Yidi Miao, Ramayya Krishnan, and Rema Padman.
\newblock {JEV}-as-a-{Judge}: Accept when confident, escalate when unsure, 2026.
\newblock arXiv:2609.26550v1, 22 September 2026.

\bibitem[Liu et~al.(2024)Liu, Zeng, Liu, Yan, He, Wang, Yan, Liu, and Zhou]{liu2024skywork}
Chris~Yuhao Liu, Liang Zeng, Jiacai Liu, Rui Yan, Jujie He, Chaojie Wang, Shuicheng Yan, Yang Liu, and Yahui Zhou.
\newblock Skywork-reward: Bag of tricks for reward modeling in {LLM}s.
\newblock \emph{arXiv preprint arXiv:2410.18451}, 2024.
\newblock Source of the \texttt{Skywork-Reward-Preference-80K-v0.2} training pairs.

\bibitem[Liu et~al.(2025{\natexlab{a}})Liu, Zeng, Xiao, He, Liu, Wang, Yan, Shen, Zhang, Xu, Liu, and Zhou]{liu2025skyworkv2}
Chris~Yuhao Liu, Liang Zeng, Yuzhen Xiao, Jujie He, Jiacai Liu, Chaojie Wang, Rui Yan, Wei Shen, Fuxiang Zhang, Jiacheng Xu, Yang Liu, and Yahui Zhou.
\newblock Skywork-reward-v2: Scaling preference data curation via human-ai synergy.
\newblock \emph{arXiv preprint arXiv:2507.01352}, 2025{\natexlab{a}}.

\bibitem[Liu et~al.(2025{\natexlab{b}})Liu, Yao, Min, Cao, Hou, and Li]{liu2025rmbench}
Yantao Liu, Zijun Yao, Rui Min, Yixin Cao, Lei Hou, and Juanzi Li.
\newblock {RM}-bench: Benchmarking reward models of language models with subtlety and style.
\newblock In \emph{International Conference on Learning Representations (ICLR)}, 2025{\natexlab{b}}.
\newblock arXiv:2410.16184.

\bibitem[Malik et~al.(2026)Malik, Pyatkin, Land, Morrison, Smith, Hajishirzi, and Lambert]{malik2025rewardbench2}
Saumya Malik, Valentina Pyatkin, Sander Land, Jacob Morrison, Noah~A. Smith, Hannaneh Hajishirzi, and Nathan Lambert.
\newblock Rewardbench 2: Advancing reward model evaluation.
\newblock In \emph{International Conference on Learning Representations (ICLR)}, 2026.
\newblock arXiv:2506.01937.

\bibitem[Mozannar and Sontag(2020)]{mozannar2020defer}
Hussein Mozannar and David Sontag.
\newblock Consistent estimators for learning to defer to an expert.
\newblock In \emph{International Conference on Machine Learning (ICML)}, 2020.

\bibitem[Naeini et~al.(2015)Naeini, Cooper, and Hauskrecht]{naeini2015bayesian}
Mahdi~Pakdaman Naeini, Gregory~F. Cooper, and Milos Hauskrecht.
\newblock Obtaining well calibrated probabilities using bayesian binning.
\newblock In \emph{Proceedings of the AAAI Conference on Artificial Intelligence}, 2015.

\bibitem[Nandakishore(2026)]{nandakishore2026laya}
Gowthamkumar Nandakishore.
\newblock Laya as a typed probabilistic assessor: An independent reproduction and a preregistered study of calibration and selective escalation, 2026.
\newblock arXiv:2609.33843.

\bibitem[Nixon et~al.(2019)Nixon, Dusenberry, Jerfel, Nguyen, Liu, Zhang, and Tran]{nixon2019measuring}
Jeremy Nixon, Mike Dusenberry, Ghassen Jerfel, Timothy Nguyen, Jeremiah Liu, Linchuan Zhang, and Dustin Tran.
\newblock Measuring calibration in deep learning.
\newblock In \emph{CVPR Workshops}, 2019.

\bibitem[Nosek et~al.(2018)Nosek, Ebersole, DeHaven, and Mellor]{nosek2018prereg}
Brian~A. Nosek, Charles~R. Ebersole, Alexander~C. DeHaven, and David~T. Mellor.
\newblock The preregistration revolution.
\newblock \emph{Proceedings of the National Academy of Sciences}, 115\penalty0 (11):\penalty0 2600--2606, 2018.

\bibitem[Ong et~al.(2024)Ong, Almahairi, Wu, Chiang, Wu, Gonzalez, Kadous, and Stoica]{ong2024routellm}
Isaac Ong, Amjad Almahairi, Vincent Wu, Wei-Lin Chiang, Tianhao Wu, Joseph~E. Gonzalez, M.~Waleed Kadous, and Ion Stoica.
\newblock Routellm: Learning to route {LLM}s with preference data, 2024.

\bibitem[Panickssery et~al.(2024)Panickssery, Bowman, and Feng]{panickssery2024own}
Arjun Panickssery, Samuel~R. Bowman, and Shi Feng.
\newblock {LLM} evaluators recognize and favor their own generations.
\newblock In \emph{Advances in Neural Information Processing Systems (NeurIPS)}, 2024.
\newblock arXiv:2404.13076.

\bibitem[{Qwen Team}(2025)]{qwen3report}
{Qwen Team}.
\newblock Qwen3 technical report, 2025.

\bibitem[Roelofs et~al.(2022)Roelofs, Cain, Shlens, and Mozer]{roelofs2022mitigating}
Rebecca Roelofs, Nicholas Cain, Jonathon Shlens, and Michael~C. Mozer.
\newblock Mitigating bias in calibration error estimation.
\newblock In \emph{International Conference on Artificial Intelligence and Statistics (AISTATS)}, 2022.

\bibitem[Tan et~al.(2025)Tan, Zhuang, Montgomery, Tang, Cuadron, Wang, Popa, and Stoica]{tan2025judgebench}
Sijun Tan, Siyuan Zhuang, Kyle Montgomery, William~Y. Tang, Alejandro Cuadron, Chenguang Wang, Raluca~Ada Popa, and Ion Stoica.
\newblock Judgebench: A benchmark for evaluating {LLM}-based judges.
\newblock In \emph{International Conference on Learning Representations (ICLR)}, 2025.
\newblock arXiv:2410.12784.

\bibitem[Wang et~al.(2024)Wang, Li, Chen, Cai, Zhu, Lin, Cao, Kong, Liu, Liu, and Sui]{wang2024fair}
Peiyi Wang, Lei Li, Liang Chen, Zefan Cai, Dawei Zhu, Binghuai Lin, Yunbo Cao, Lingpeng Kong, Qi~Liu, Tianyu Liu, and Zhifang Sui.
\newblock Large language models are not fair evaluators.
\newblock In \emph{Proceedings of the 62nd Annual Meeting of the Association for Computational Linguistics (ACL)}, 2024.
\newblock arXiv:2305.17926.

\bibitem[Wilson(1927)]{wilson1927probable}
Edwin~B. Wilson.
\newblock Probable inference, the law of succession, and statistical inference.
\newblock \emph{Journal of the American Statistical Association}, 22\penalty0 (158):\penalty0 209--212, 1927.

\bibitem[Zadrozny and Elkan(2002)]{zadrozny2002isotonic}
Bianca Zadrozny and Charles Elkan.
\newblock Transforming classifier scores into accurate multiclass probability estimates.
\newblock \emph{Proceedings of the Eighth ACM SIGKDD International Conference on Knowledge Discovery and Data Mining}, 2002.

\bibitem[Zheng et~al.(2023)Zheng, Chiang, Sheng, Zhuang, Wu, Zhuang, Lin, Li, Li, Xing, Zhang, Gonzalez, and Stoica]{zheng2023mtbench}
Lianmin Zheng, Wei-Lin Chiang, Ying Sheng, Siyuan Zhuang, Zhanghao Wu, Yonghao Zhuang, Zi~Lin, Zhuohan Li, Dacheng Li, Eric~P. Xing, Hao Zhang, Joseph~E. Gonzalez, and Ion Stoica.
\newblock Judging {LLM}-as-a-judge with {MT}-bench and chatbot arena.
\newblock In \emph{Advances in Neural Information Processing Systems (NeurIPS)}, 2023.
\newblock arXiv:2306.05685.

\end{thebibliography}

\end{document}